\pdfoutput=1

\documentclass[11pt]{article}

\usepackage[]{acl}
\usepackage{times}
\usepackage{latexsym}

\usepackage[normalem]{ulem}
\usepackage{mathrsfs}
\usepackage{amsthm}
\usepackage{amsmath}
\usepackage{amssymb, bm}
\usepackage{multirow}
\usepackage{booktabs}
\usepackage{enumitem}
\usepackage{times}
\usepackage{latexsym}
\usepackage{multirow}
\usepackage{graphicx}
\usepackage{amsfonts}
\usepackage{subcaption}
\usepackage{xspace}
\usepackage{tabularx}
\usepackage{algorithm}
\usepackage{arydshln}
\usepackage{algpseudocode}

\newcommand{\ourmodel}{TASA\xspace}
\newcommand{\secref}[1]{\S\ref{#1}}
\algnewcommand\algorithmicinput{\textbf{Input:}}
\algnewcommand\algorithmicoutput{\textbf{Output:}}
\algnewcommand\algorithmicdef{\textbf{Definition:}}
\algnewcommand\Input{\item[\algorithmicinput]}%
\algnewcommand\Output{\item[\algorithmicoutput]}%
\algnewcommand\Def{\item[\algorithmicdef]}%

\definecolor{mypurple}{RGB}{111,61,121}
\definecolor{myblue}{RGB}{46,88,180}
\definecolor{myred}{RGB}{181,68,106}
\definecolor{textorange}{RGB}{237,125,49}
\definecolor{textblue}{RGB}{46,117,181}
\definecolor{textgreen}{RGB}{112,173,71}

\newcommand{\setorange}[1]{{\color{textorange}{#1}}}
\newcommand{\setblue}[1]{{\color{textblue}{#1}}}
\newcommand{\setgreen}[1]{{\color{textgreen}{#1}}}
\newcommand{\setred}[1]{{\color{myred}{#1}}}

\usepackage[T1]{fontenc}

\usepackage[utf8]{inputenc}

\usepackage{microtype}

\title{TASA: Deceiving Question Answering Models by\\ Twin Answer Sentences Attack}

\author{Yu Cao\textsuperscript{1}\thanks{\;\;Work was done when Yu Cao was an intern at JD Explore Academy.}\;, Dianqi Li\textsuperscript{2}, Meng Fang\textsuperscript{3}, Tianyi Zhou\textsuperscript{4}, Jun Gao\textsuperscript{5}, Yibing Zhan\textsuperscript{6}, Dacheng Tao\textsuperscript{1} \\
\textsuperscript{1}School of Computer Science, The University of Sydney \\
\textsuperscript{2}University of Washington\quad \textsuperscript{3}University of Liverpool\quad  \textsuperscript{4}University of Maryland\\
\textsuperscript{5}Harbin Institute of Technology, Shenzhen\quad \textsuperscript{6}JD Explore Academy\\
\tt ycao8647@uni.sydney.edu.au, dianqili@uw.edu \\ 
\tt Meng.Fang@liverpool.ac.uk, zhou@umiacs.umd.edu\\
\tt jgao95@stu.hit.edu.cn, zhanyibing@jd.com,  dacheng.tao@gmail.com
}

\begin{document}

\maketitle

\begin{abstract}
We present \textbf{T}win \textbf{A}nswer \textbf{S}entences \textbf{A}ttack (\ourmodel), an adversarial attack method for question answering (QA) models that produces fluent and grammatical adversarial contexts while maintaining gold answers. Despite phenomenal progress on general adversarial attacks, few works have investigated the vulnerability and attack specifically for QA models. In this work, we first explore the biases in the existing models and discover that they mainly rely on keyword matching between the question and context, and ignore the relevant contextual relations for answer prediction.
Based on two biases above, \ourmodel attacks the target model in two folds: (1) lowering the model's confidence on the gold answer with a \emph{perturbed answer sentence}; 
(2) misguiding the model towards a wrong answer with a \emph{distracting answer sentence}. Equipped with designed beam search and filtering methods, \ourmodel can generate more effective attacks than existing textual attack methods while sustaining the quality of contexts, in extensive experiments on five QA datasets and human evaluations.
\end{abstract}

\section{Introduction}
Question Answering (QA) is the cornerstone of various NLP tasks. In extractive QA (the most common setting), given a question and an associated context, a QA model needs to comprehend on the context and predict the answer~\cite{squad1.1}. While most works keep improving the answer correctness on benchmarks~\cite{bert,qanet}, few studies investigate the robustness of QA models, e.g., is the performance achieved by sound contextual comprehension or via shortcuts like keyword matching? Although adversarial attacks attract growing interests in computer vision~\cite{fgsm,gan_adversarial} and recently in NLP~\cite{pwws,clare}, most of them study general tasks without taking into account the properties of QA. The vulnerability and biases of models can lead to catastrophic failures outside the benchmark datasets. An effective way to study them is through adversarial attacks specifically designed for QA tasks.  

Generating adversarial textual examples is challenging due to the discrete syntactic restriction, especially on QA, where the additional relationship between question and context should be further considered.
Existing works such as AddSent and Human-in-the-loop~\cite{addsent,human_in_loop} heavily rely on human annotators to create effective adversarial QA examples, which are costly and hard to scale. A few studies~\cite{question_paraphrase,t3,universal_trigger} can generate adversarial samples automatically. But they only perturb either the context or the question separately, and thus ignore the consistency between them. Moreover, the major pitfalls of QA models' detailed comprehension process are not fully investigated, confining producing more powerful adversarial attacks.

\begin{figure*}[!t]
    \setlength{\abovecaptionskip}{0.1cm}
    \setlength{\belowcaptionskip}{-0.2cm}
    \centering
    \includegraphics[width=.98\linewidth]{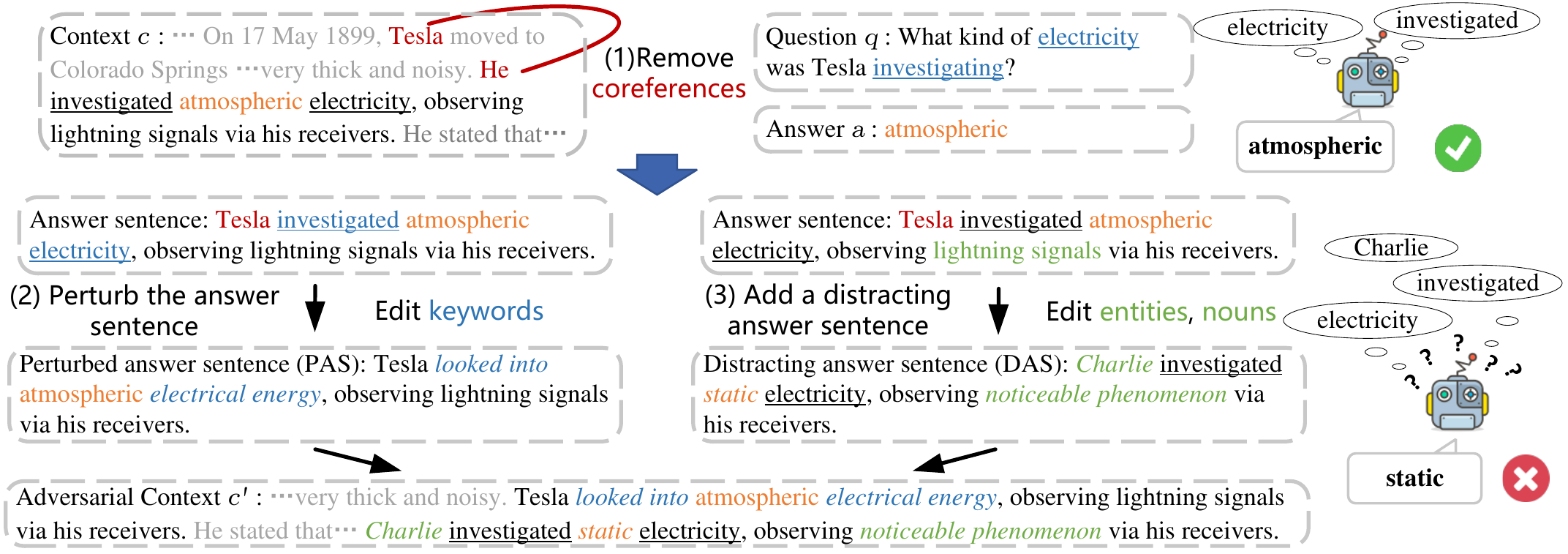}
    \caption{An example of \ourmodel generating adversarial context $C'$. \underline{Underlined} parts indicate keywords. Orange indicates \setorange{gold answer} or \setorange{\emph{pseudo answer}}. Other colors indicate tokens for \setblue{perturbation}, \setgreen{distracting}, or \textcolor[RGB]{192,0,0}{coreferences}. }
    \label{fig:main_example}
\end{figure*}

In this paper, we develop an adversarial attack specifically targeting two biases of mainstream QA models discussed in \secref{sec:prove_experiment}: (1) making prediction via keywords matching in the answer sentence of contexts; and (2) ignorance of the entities shared between the question and context. Our method, \textbf{T}win \textbf{A}nswer \textbf{S}entences \textbf{A}ttack (\ourmodel), automatically produces black-box adversarial attacks~\cite{black_box} perturbing a context without hurting its fluency or changing the gold answer. 
\ourmodel firstly allocates the answer sentence in the context that is decisive for answering~\cite{qa_extract_sentence} and then modify it into two sentences targeting the two biases above: one sentence preserves the gold answer and the meaning but replaces the keywords that are shared with the question with their synonyms; while the other leaves the keywords and the syntactic structure intact but changes the entities (subjects/objects) associated with the answer. Thereby, the former is a {\it perturbed answer sentence} (PAS) lowering the focus of the model on the gold answer, while the latter generates a {\it distracting answer sentence} (DAS) as~\citet{addsent} to further misguide the model towards a wrong answer with respect to irrelevant entities. Thus, the adversarial context can substantially distort the QA model without changing the answer for humans. To address the challenge of efficiency and textual fluency, we further propose beam search and filtering techniques empowered by pretrained models.

In experiments, we evaluate \ourmodel and other adversarial attack baselines on attacking three popular contextualized QA models, BERT~\cite{bert}, SpanBERT~\cite{spanbert}, and BiDAF~\cite{bidaf}, on five extractive QA datasets, i.e., SQuAD 1.1, NewsQA, NaturalQuestions, HotpotQA, and TriviaQA. Experimental results and human evaluations consistently show that \ourmodel achieves better attack capability than other baselines and meanwhile preserves the textual quality and gold answers identifiable by humans. 

Our contributions are three-fold:
\begin{itemize}[wide=1\parindent, noitemsep, topsep=0pt]
    \item We propose a novel adversarial attack method ``\ourmodel'' specifically designed to fool extractive QA models while retain the gold answers for humans. 
    \item We study the biases and vulnerability of QA models that motivate \ourmodel, and demonstrate that those models mainly rely on keyword matching, while may ignore the contextual relation.
    \item Experiments on five QA benchmark datasets and three types of victim models demonstrate that \ourmodel outperforms existing baselines on attack performance, as well as the comparable capability to preserve textual quality and answers.
\end{itemize}

We release our code at \url{https://github.com/caoyu-noob/TASA}

\section{\label{sec:prove_experiment}Predicting Bias in Question Answering}

Recent works show that state-of-the-art natural language inference models often overly rely on certain keywords as shortcuts for prediction~\cite{universal_trigger,unnatural_nli}. In the empirical study of this section, we illustrate that current QA models consistently exhibit such bias on the sensitive words without leveraging the contextual relationship for predicting answers. 


We analyze two mainstream QA models with contextualized comprehension capabilities, BERT~\cite{bert} and BiDAF~\cite{bidaf}, trained on the original training set of SQuAD1.1~\cite{squad1.1} and tested on samples modified from its validation set. We define the sentence in the context that contains the gold answer as the \textbf{answer sentence}, which is the key for predicting answers~\cite{qa_extract_sentence}. We first compare the performance of models on the original sample with only answer sentence as the context (``\emph{Only answer sent.}'').
Besides, to investigate the bias on sensitive words, we further examine models on samples with various types of sensitive words in the answer sentence being (1) either removed (``\emph{Remove}'') or (2) only retained (``\emph{Only}'').
There are three types of sensitive words to be considered:

\noindent \textbf{(1) Entities.} The same named entities shared between the answer sentence and the question.

\noindent \textbf{(2) Lexical words~(lexical.).} with lexical meanings (excluding all named entities) shared between the answer sentence and question. They cover the words with POS tags of \textit{NOUN}, \textit{VERB}, \textit{ADJ}, etc.\looseness-1

\noindent \textbf{(3) Function words~(func.).} Words that do not have lexical meaning but are shared between the answer sentence and the question. They include words with POS tags of \textit{DET}, \textit{ADP}, \textit{PRON}, etc.\looseness-1

\begin{figure}[!t]
    \setlength{\belowcaptionskip}{-0.2cm}
    \centering
    \includegraphics[width=1\linewidth]{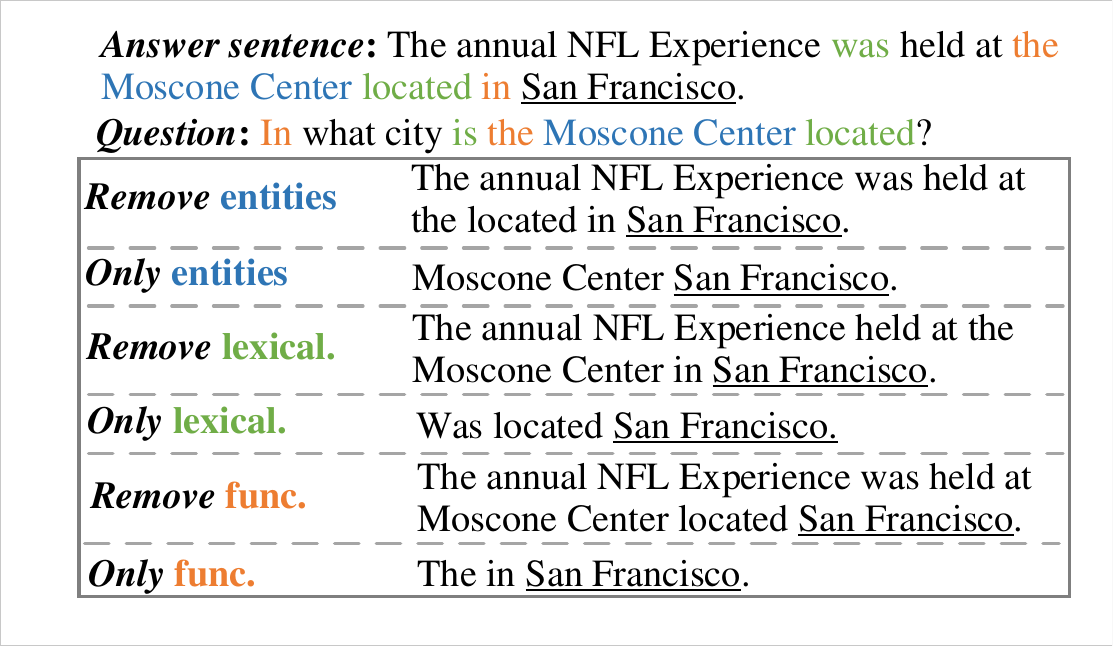}
    \caption{The illustration of removing or only retaining (\emph{Only}) different types of sensitive words, the answer is \underline{underlined} and kept. }
   \label{fig:prove_example}
\end{figure}

When modifying the answer sentence, we only remove or retain these three types of sensitive words, except the gold answer words, and also keep the rest context intact. As shown in Figure~\ref{fig:prove_example}, the modified texts are unreadable and difficult to infer their true meaning from the human perspective. In addition, we follow UNLI~\cite{unnatural_nli} to \textbf{Shuffle} tokens in the answer sentence for \emph{Only} lexical. conditions, verifying the possibility of models to achieve even better performance, given the texts are totally ungrammatical but contain sensitive words.

\begin{table}[]
\setlength{\tabcolsep}{1.2mm}
\setlength{\belowcaptionskip}{-0.2cm}
\small
\centering
\begin{tabular}{lrrrr}
\toprule
Model & \multicolumn{2}{c}{BERT} & \multicolumn{2}{c}{BiDAF} \\
\midrule
& EM & F1 & EM & F1 \\
\midrule
Original & 80.91 & 88.23 & 65.72 & 75.97 \\
\emph{Only answer sent.} & +2.79 & +2.87 & +3.27 & +4.37 \\
\midrule
\emph{Remove} entities & -5.39 & -4.17 & -4.84 & -6.27 \\
\emph{Only} entities & -23.42 & -15.90 & -26.75 & -18.03 \\
\midrule
\emph{Remove} lexical. & -18.62 & -16.71 & -24.43 & -24.46 \\
\emph{Only} lexical. & -5.27 & -1.81 & +0.86 & +4.28 \\
\emph{Only} lexical. (shuffle) & +4.07 & +2.94 & +10.68 & +10.46 \\
\midrule
\emph{Remove} func. & -5.20 & -3.18 & -5.42 & -3.55 \\
\emph{Only} func.  & -24.08 & -22.34 & -22.24 & -22.72 \\
\bottomrule
\end{tabular}
\caption{\label{tab:prove_results} EM and F1 scores of BERT and BiDAF models on different modified samples compared to results on the original samples. Shuffle means the best results among texts whose tokens are random-ordered.}
\end{table}


Table~\ref{tab:prove_results} compares the evaluation results on different modifications. Given the answer-sentence-only context, the performance of both BERT and BiDAF are improved, indicating that they mainly rely on the answer sentence and almost ignore the rest of the context.
While removing entities or function words causes a slight difference in metrics, removing lexical words leads to a larger performance drop. In addition, both models perform surprisingly satisfactory when keeping only lexical words in answer sentences, compared to the 30\% $\sim$ 60\% drop when keeping other words. Moreover, shuffling tokens under the lexical-only conditions even possibly benefit the model despite the answer sentence being merely discrete tokens and hard to read. This suggests that both models can answer questions solely relying on the shared lexical words (not contextual), i.e., \emph{keywords} in the answer sentence, regardless of the word order and other contextual information like entities. \looseness-1 

Inspired by this observation, we question that whether we can utilize the discovered pitfall to design an efficient adversarial attack method specifically for QA?
Can we lower the model's attention on the gold answers and then misguide it to incorrect answers by manipulating the existing sensitive keywords in the context and adding some new misleading ones? The answer is affirmative: we show that the predictions can be shifted to crafted wrong answers in~\secref{sec:more_analysis}.\looseness-1 

\section{Methodology}

We propose an adversarial attack method for extractive QA models, \textbf{T}win \textbf{A}nswer \textbf{S}entences \textbf{A}ttack (\ourmodel), which automatically produces black-box attacks solely based on the final output of the \textbf{victim QA model} $F(\cdot)$. Given a typical QA sample composed of a \textbf{context} $c$, a \textbf{question} $q$, and an \textbf{answer} $a$ (i.e., a positional span in $c$), we study how to \textbf{perturb the context} $c$ as $c'$ that can deceive $F(\cdot)$ towards producing an incorrect answer $F(c',q) \neq a$, while $c'$ retains the correct answer $a$ that can be identified by humans.
We keep the question $q$ intact to ensure the answer $a$ valid, as editing the short $q$ with simple syntactic structure easily alters its meaning.

\ourmodel can be summarized as three main steps: (1) Remove coreferences in the context to facilitate the following edits; (2) Perturb the \emph{answer sentence} by replacing \emph{keywords} (overlapped sensitive lexical words in \secref{sec:prove_experiment}) with synonyms to produce a {\it perturbed answer sentence} (PAS), lowering the model's focus on the gold answer; (3) Add a {\it distracting answer sentence} (DAS) that keeps the \emph{keywords} intact but changes the associated subjects/objects to misguide the model for producing a wrong answer, which can be proven in Table~\ref{tab:pseudo_answer}. 
How these the three steps are applied is illustrated in Figure~\ref{fig:main_example}. And Algorithm~\ref{alg:tasa} gives the complete procedure of \ourmodel.

\subsection{Remove Coreferences}

Coreference relations across sentences commonly exist in texts~\cite{coreference} and also bring extra challenges to adversarial attacks during making substitutions on target words.
For example, in a sentence ``\textit{His patented AC induction motor were licensed}'', ``\textit{His}'' refers to ``\textit{Nikola Tesla's}''  according to the whole context. However, given the single sentence, it is hard to precisely allocate candidates for substitution ``\textit{his}'' as it is a pronoun. 
Instead, we remove the coreference by replacing such pronouns with the entity names they refer to, e.g., specific persons or locations, so we can edit them directly without considering a complicated coreference.


\subsection{\label{sec:PAS}Perturb the Answer Sentence}

According to the former analysis, the \emph{answer sentence} is the most important part of context $c$ for QA tasks, and QA models usually predict answers according to keyword matching~\cite{qa_extract_sentence}.
Hence, we first study how to obtain a perturbed answer sentence (PAS) by only perturbing those sensitive \emph{keywords} instead of changing the whole context.
Given the gold answer $a$, we first allocate the \emph{answer sentence} $s_a$ in $c$. In \ourmodel, we use the text matching to search for $s_a$ that contains text $a$.\looseness-1

\noindent \textbf{Determine the keywords to perturb.} As discussed in~\secref{sec:prove_experiment}, QA models normally rely on \emph{keywords} to make predictions. Hence, we directly perturb those keywords rather than randomly-selected tokens as previous works~\cite{pwws,textfooler} to produce more effective attacks. 
We adopt three criteria to select words of $s_a$ into the keyword set $\mathcal{X}$: (1) they are not included in the answer span $a$ so the gold answer will retain; and (2) each of them shares the same lemma with a token in the question $q$; and (3) each keyword's POS tag belongs to a POS tag set for lexical words, e.g., \textit{NOUN}, \textit{ADJ}, etc.

\noindent \textbf{Rank keywords by importance.} Following previous works~\cite{textfooler}, we rank keywords in $\mathcal{X}$ according to their importance scores in the descending order. Given the original context $c$ and answer $a$, the importance score $I_i$ of $x_i \in \mathcal{X}$ is
\begin{eqnarray}
\label{eq:importance}
    I_i = p_F(a|c,q) - p_F(a|mask(c, x_i),q),
\end{eqnarray}
where $p_F(a|\cdot)$ denotes the probability of the original span position of gold answer $a$ predicted by the victim model $F$,
$mask(c, x_i)$ means $c$ is modified by replacing a token $x_i$ with a special mask symbol, e.g., given $c=$ $..x_{i-1}x_ix_{i+1}..$,  $mask(c, x_i)=..x_{i-1}<mask>x_{i+1}..$. Finally, we obtain a set $\mathcal{X}$ of keywords ranked by their importance.

\noindent \textbf{Generate perturbed answer sentence (PAS).} Following the order in $\mathcal{X}$, we edit each keyword $x_i \in \mathcal{X}$ one after another. 
Specifically, we replace $x_i$ with its synonym $r_j$ from a synonym set and transform the inflection of $r_j$ as the same as $x_i$, e.g., we change ``\textit{Tesla investigated...}'' to ``\textit{Tesla looked into...}'' where ``\textit{investigated}'' is a keyword and ``\textit{look into}'' is one of its synonyms.

Thereby, multiple PASs are obtained during editing each keyword if more than one synonym exists. We retain the top few of them via beam search and filtering strategy (as elaborated in~\secref{sec:filter}) to promote the effectiveness as well as efficiency, resulting in a set of PASs $\mathcal{P}$, which will be the initial texts of the next perturbation turn. While PASs do not change the meaning of texts as they replace words with their synonyms, they will distract the model, which relies on keyword matching, away from PAS containing the answer.


\subsection{Add a Distracting Answer Sentence}

To further deceive the model, we also add a distracting answer sentence (DAS) at the end of the context.
In particular, DAS is modified from the \emph{answer sentence} $s_a$ as well: it changes the subjects/objects and the answer, but keeps the keywords intact which can draw models' attention. 
Collaborating with PAS, DAS misguides models to predict incorrect answers regarding wrong subjects/objects due to the pitfall studied in~\secref{sec:prove_experiment}, which will be verified in Table~\ref{tab:pseudo_answer}.
Our method differs from previous works~\cite{addsent,universal_trigger} as our distractors are added automatically and suits more general conditions. 

\noindent \textbf{Determine the tokens to edit.} Similar to PAS, the first step of generating DAS is to select a set $\mathcal{Y}$ of tokens from the $s_a$ as the candidates of subjects/objects that will be edited.
In \ourmodel, each selected token $y\in\mathcal{Y}$ needs to meet all the following criteria: (1) $y\notin \mathcal{X}$ so the original keywords are preserved; (2) $y\notin a$ (as we will process the answer tokens separately); (3) $y$ is a named entity or its POS tag is NOUN. 
The goal of (3) is to extract and change the subjects/objects of $s_a$ to produce a pseudo answer sentence that contains incorrect answers.  
We do not use a syntactic parser to locate the subjects/objects, as we find it less accurate and effective than POS tags empirically. 

\noindent \textbf{Generate distracting answer sentence (DAS).} Similar to PAS, we edit each $y_i \in \mathcal{Y}$ to obtain a DAS. Specifically, we replace each $y_i$ with a token/phrase of the same entity/noun type, e.g., ``\textit{Tesla investigated...}'' can be modified to ``\textit{Charlie investigated...}'' since both ``\textit{Tesla}'' and ``\textit{Charlie}'' are persons. In principle, (1) if $y_i$ is a named entity, we randomly sample $N$ different entities with the same NER tag from the whole corpus as the candidates to replace $y_i$; (2) otherwise, we randomly sample $N$ nouns with the same hypernym as $y_i$ from the corpus for substitution. Hence, multiple DASs can be generated, and we also use the beam search strategy to only choose the top few of them, resulting in a set of DASs $\mathcal{D}$.


\noindent \textbf{Change the answer in DAS.} 
Since the main purpose of DAS is to misguide the model to predict a wrong answer, we entirely replace the text span of the original answer in DAS with a pseudo answer, which helps to remove the ambiguity of the answer from humans' perspective.
Specifically, we replace every lexical token of $a$ in DAS with one of pseudo answer token candidates that share the same NER tags or POS tags, which are randomly sampled from the whole corpus. Likewise, this procedure results in multiple results and thus a beam search is also necessary for the efficiency and attack success purpose as well.

\begin{algorithm}[ht!]
\small
\caption{TASA}
\label{alg:tasa}
\begin{algorithmic}[1]
    \Def Beam size $M$, importance score $I_i$ given in Eq.~\ref{eq:importance}, effect score $E_n$ given in Eq.~\ref{eq:effect_score}, threshold $T_E$ for $E_n$
    \Input QA sample $(c,q,a)$, victim model $F(\cdot)$
    \Output An adversarial context $c'$ to fool $F(\cdot)$
    \State Remove coreferences in $c$;
    \State Extract answer sentence $s_a$ from $c$;
    \State $\mathcal{X} \gets$ keywords in $s_a$ and rank them by $I_i$;
    \State \textbf{Initialize the PAS set:} $\mathcal{P} \gets \{s_a\}$
    \For{$1 \leq i \leq |\mathcal{X}|$}
        \State $\mathcal{P} \gets$ perturb $x_i$ for each item in $\mathcal{P}$;
        \State $\mathcal{P} \gets$ $M$ items in $\mathcal{P}$ with the highest $E_n$;
        \If{$T_E \leq$ minimum $E_n$ in $\mathcal{P}$} \textbf{break;}
        \EndIf
    \EndFor
    \State $\mathcal{P} \gets$ filter $\mathcal{P}$ based on answerable and quality in \secref{sec:filter};
    \State \textbf{Initialize the DAS set:} $\mathcal{D} \gets \{(s_j, c_j)\}$, each DAS $s_j=s_a$, paired with context $c_j$ modified by each PAS in $\mathcal{P}$;
    \State $\mathcal{Y} \gets$ a set of tokens in $s_a$ to be edited for DAS;
    \For{$1 \leq i \leq |\mathcal{Y}|$}
        \State $\mathcal{D} \gets$ edit $y_i$ for each DAS $s_j$ in $\mathcal{D}$;
        \State $\mathcal{D}\gets$ $M$ items in $\mathcal{D}$  with the highest $E_n$;
    \EndFor
    \State  $\mathcal{D}\gets$ edit answer tokens for each DAS $s_j$ in $\mathcal{D}$;
    \State $(s_b, c_b) \gets$ The item in $\mathcal{D}$ with the highest $E_n$;
    \State $c' \gets$ append DAS $s_b$ to the end of context $c_b$;
    \State \textbf{return} $c'$;
\end{algorithmic}
\end{algorithm}

\subsection{\label{sec:filter}Beam Search and Filtering}

\noindent \textbf{Beam search.} When editing each word in generating the PAS and DAS, there usually exist multiple replacement candidates, resulting in multiple perturbed sentences. 
In order to obtain the one that has the greatest potential leading to a successful attack, and to improve the attack's efficiency, we apply a beam search strategy defined based on the effect score $E_n$ for each perturbed sentence $s_n$. 
\begin{eqnarray}
\label{eq:effect_score}
    E_n = p_F(a|c,q) - p_F(a|edit(c, s_n),q),
\end{eqnarray}
where $edit$ denotes that the original context $c$ is modified by $s_n$: (1) if $s_n$ is a PAS, it replaces the original answer sentence $s_a$ in $c$; (2) if $s_n$ is a DAS, it is appended to the end of $c$. These edited texts will be ranked by $E_n$ in the descending order, and only the top $M$ ($M$ is beam size) are retained for the next edit step. 
Beam search will stop if (1) no additional edit is needed for the current sample,
or (2) the minimum effect score among the result is higher than a threshold $T_E$ that can ensure sufficient performance drop. 

\ourmodel runs beam search for PAS to obtain a PAS set $\mathcal{P}$, then obtain a DAS set $\mathcal{D}$ sequentially, and finally generate the adversarial context $c'$. Note that we obtain a DAS based on a series of contexts that are already perturbed by $\mathcal{P}$. So each item in $\mathcal{D}$ is a pair of a DAS $s_j$ and a corresponding perturbed context $c_j$, and the initial $\mathcal{D}$ contains all possible contexts edited by each PAS in $\mathcal{P}$.

\noindent \textbf{Filtering by textual quality.} To ensure high textual quality and answer preservation
of the generated adversarial contexts, \ourmodel applies a filtering procedure on the $M$ (beam size) PASs
achieved after the final beam search for generating PAS. We skip it for DASs as they have no effect on the gold answer.
In particular, we firstly use a model to justify whether the question $q$ is still answerable given the perturbed context $edit(c, s_n)$. 
Such a model can be a large-scale pretrain model fine-tuned on both answerable and unanswerable samples (refer to Appendix~\ref{app:implementation} for details). Only those contexts classified as \emph{answerable} will remain. 
In addition, we further constrain the remained contexts' textual quality in terms of semantic similarity and fluency:
\begin{eqnarray}
\label{eq:quality_index}
    U_n = USE(s_n, s_a) - PPL(s_n) / PPL(s_a),
\end{eqnarray}
where $USE$ denotes the USE similarity~\cite{use} between two sentences and $PPL$ denotes the perplexity computed by a GPT2 model~\cite{gpt2}. Only $s_n$ fulfilling $U_n \geq T_U$ ($T_U$ as a threshold) are retained for the next step.

\section{Experiments}
\label{sec:exp}

\begin{table*}[t!]
\setlength{\abovecaptionskip}{0.1cm}
\setlength{\belowcaptionskip}{-0.2cm}
\small
\centering
\begin{tabular}{ll|rrrrr|rrrrr}
\toprule
\multicolumn{2}{l|}{\textbf{Victim model}} &  \multicolumn{5}{c|}{\textbf{BERT-base}} & \multicolumn{5}{c}{\textbf{SpanBERT-large}} \\
\midrule
\textbf{Dataset} & \textbf{method} & \textbf{EM$\downarrow$} & \textbf{F1$\downarrow$} & \textbf{GErr$\downarrow$} & \textbf{PPL$\downarrow$} & \textbf{Num} & \textbf{EM$\downarrow$} & \textbf{F1$\downarrow$} & \textbf{GErr$\downarrow$} & \textbf{PPL$\downarrow$} & \textbf{Num} \\
\midrule
\multirow{4}{0pt}{SQuAD 1.1} & Original & 80.91 & 88.23 & 2.39 & 33.25 & 10,570 & 88.25 & 94.00 & 2.39 & 33.25 & 10,570 \\
& AddSent* & 57.78 & 64.58 & 2.47 & 33.98 & 3,560 & 73.88 & 79.77 & 2.47 & 33.98 & 3,560 \\
& TextFooler & 67.18 & 78.18 & \textbf{2.95} & 44.84 & 7,919 & 80.00 & 88.08 & \textbf{3.03} & 44.57 & 7,746 \\
& T3 & 71.63 & 78.86 & 3.48 & 44.45 & 9,622 & 76.76 & 82.13 & 3.66 & 42.30 & 9,761 \\
& OURS & \textbf{40.06} & \textbf{50.87} & 2.98 & \textbf{41.15} & 9,559 & \textbf{54.18} & \textbf{65.50} & 3.09 & \textbf{40.31} & 9,580 \\
\midrule
\multirow{4}{0pt}{NewsQA} & Original & 51.57 & 65.57 & 1.98 & 22.50 & 4,212 & 58.78 & 73.81 & 1.98 & 22.50 & 4,212 \\
& TextFooler & 43.31 & 58.34 & \textbf{2.14} & 24.33 & 3,727 & 52.13 & 68.25 & \textbf{2.16} & 24.80 & 3,685 \\
& T3 & \textbf{39.54} & 53.49 & 2.33 & \textbf{22.86} & 3,865 & 51.29 & 66.40 & 2.23 & 22.89 & 3,875 \\
& OURS & 39.62 & \textbf{53.46} & 2.16 & \textbf{22.86} & 2,860 & \textbf{49.96} & \textbf{64.93} & 2.18 & \textbf{22.77} & 2,872 \\
\midrule
\multirow{4}{0pt}{NQ} & Original & 67.39 & 79.28 & 20.48 & 49.74 & 12,836 & 71.74 & 83.12 & 20.48 & 49.74 & 12,836 \\
& TextFooler & 48.31 & 63.08 & 20.46 & 49.02 & 7,158 & 55.47 & 69.49 & \textbf{20.37} & 45.67 & 7,252 \\
& T3 & 60.06 & 71.20 & 20.93 & 60.90 & 10,439 & 57.92 & 70.03 & 20.78 & 63.21 & 10,446 \\
& OURS & \textbf{43.23} & \textbf{55.32} & \textbf{20.42} & \textbf{44.30} & 8,809 & \textbf{51.08} & \textbf{64.84} & 20.40 & \textbf{15.24} & 8,829 \\
\midrule
\multirow{4}{0pt}{HotpotQA} & Original & 56.89 & 75.70 & 3.73 & 17.01 & 5,901 & 63.29 & 81.60 & 3.73 & 17.01 & 5,901 \\
& TextFooler & 33.59 & 47.76 & 4.01 & 20.52 & 5,397 & 60.49 & 78.94 & 4.04 & 20.96 & 5,369 \\
& T3 & 30.45 & 42.08 & 4.81 & 21.17 & 5,669 & 53.89 & 70.55 & 4.80 & 20.96 & 5,583 \\
& OURS & \textbf{27.01} & \textbf{39.10} & \textbf{3.99} & \textbf{17.29} & 5,345 & \textbf{44.18} & \textbf{60.50} & \textbf{3.99} & \textbf{17.24} & 5,355 \\
\midrule
\multirow{4}{0pt}{TriviaQA} & Original & 58.61 & 65.42 & 3.74 & 24.42 & 7,785 & 67.51 & 74.38 & 3.74 & 24.42 & 7,785 \\
& TextFooler & 52.51 & 57.39 & 4.30 & 25.85 & 7,307 & \textbf{63.62} & 69.95 & \textbf{3.81} & 25.78 & 7,358 \\
& T3 & 51.85 & 56.06 & 4.06 & \textbf{24.49} & 7,543 &  64.12 &  \textbf{69.62} & 4.07 & \textbf{24.50} & 7,549 \\
& OURS & \textbf{51.50} & \textbf{54.23} & \textbf{3.81} & 24.69 & 7,092 & 63.86 & 69.97 & 3.85 & 24.65 & 7,103 \\
\bottomrule
\end{tabular}
\caption{\label{tab:main_results}Main results on 5 QA datasets. The best results are in \textbf{bold}. \textbf{Num} is the sample number of a dataset or generated adversarial samples from the whole dataset by a method. $\downarrow$ means that the lower value is the better. *: samples are annotated by humans. }
\end{table*}

We evaluate \ourmodel on extractive QA tasks. We begin by details of setup (\secref{sec:exp_setup}), then introduce the main results in~\secref{sec:exp_main}, followed by ablation studies in \secref{sec:exp_ablation} and additional analysis in \secref{sec:more_analysis} to better illustrate each module in our method.

\subsection{Setup}
\label{sec:exp_setup}

\noindent \textbf{Datasets.} We evaluate the QA adversarial attacks generated by \ourmodel using 5 extractive QA datasets: SQuAD 1.1~\cite{squad1.1}, NewsQA~\cite{newsqa}, Natural Questions (NQ)~\cite{natural_questions}, HotpotQA~\cite{hotpotqa}, and TriviaQA~\cite{triviaqa}. We use the settings from MRQA~\cite{mrqa} for the latter four datasets, more details are given in Appendix~\ref{app:dataset}. We report results on their dev sets, as not all their test sets are publicly available.

\medskip

\noindent \textbf{Victim models.} We attack three QA models, i.e., BERT~\cite{bert}, SpanBERT~\cite{spanbert}, and BiDAF~\cite{bidaf}, in our experiments. The former two are on the top of pretrained BERT$_{\text{base}}$ and SpanBERT$_{\text{large}}$ respectively.
Both of them benefit from huge corpora, where SpanBERT can also be regarded as one of the SOTA models for general extractive QA tasks. The latter BiDAF is an end2end model based on LSTM and bidirectional attention specially designed for extractive QA (Related results are provided in Appendix~\ref{app:bidaf_results} as it is not a SOTA model). 
All models output the start and end positions of the answer span in the context as the prediction.

\medskip

\noindent \textbf{Implementation.} Given a dataset, we firstly train each kind of models on its training set to get a model achieving satisfactory performance on the dev set. The trained model is then used as a victim model $F(\cdot)$ and we perform an adversarial attack using all samples from the \textbf{whole dev} set. We use a beam size $M=5$ for TASA. The synonym set used for PAS is obtained by unionizing two sources, i.e., (1) \textbf{WordNet} synonym dictionary~\cite{wordnet} and (2) \textbf{PPDB 2.0} dataset containing token-level paraphrase pairs~\cite{ppdb_synonyms}. 
More details about \ourmodel can be found in Appendix~\ref{app:implementation}. 

\medskip

\noindent \textbf{Baselines.} We consider the following 2 strong baselines\footnote{We run the codes provided by the original papers to get results. We use the \emph{black-box} and \emph{targeted} config for T3.} besides the original dev set (Original).
\begin{itemize}[wide=1\parindent, noitemsep, topsep=0pt]
    \item \textbf{TextFooler}~\cite{textfooler}: A general token-level attack method using synonyms derived from counter-fitting word embeddings. We directly apply it to the context $c$ to make perturbations and use the model's prediction $F_a(\cdot)$ on the gold answer to determine whether to stop attacking.
    \item \textbf{T3}~\cite{t3}: A tree-autoencoder-based method to obtain perturbed sentences for attacking. It can be directly applied to QA by adding a distracting sentence to the context. 
\end{itemize}
Both of them and our TASA are \textbf{black-box} attack methods without using the internal parameters of victim models.
We also include human-annotated \textbf{AddSent} adversarial data~\cite{addsent} for SQuAD 1.1, as they share the same contexts.

\medskip

\noindent \textbf{Evaluation metrics.} Following the former works~\cite{squad1.1,t3,clare}, we evaluate attack methods using the following metrics: 1) \textbf{EM}, the exact match ratio of predicted answers; 2) \textbf{F1}, the F1 score between the predicted answers and the gold answers. Lower EM and F1 mean better attack effectiveness; 3) \textbf{Grammar error (GErr)}, the context grammatical error numbers given by LanguageTool\footnote{https://languagetool.org/} following~\citet{word_level_attack}, we use the average value per 100 tokens due to various context lengths among datasets; 4) \textbf{PPL}, the average perplexity of all adversarial contexts given by a small sized GPT2 model~\cite{gpt2} to measure their fluency~\cite{ppl_fluency}. Lower values of \textbf{GErr} and \textbf{PPL} indicate better textual quality.

\subsection{Main Results}
\label{sec:exp_main}

The main experimental results on BERT and SpanBERT are summarized in Table~\ref{tab:main_results}. \ourmodel achieves the overall best performance among all methods. In particular, it shows the best capability to deceive models than others on 3 datasets and the comparable best results on NewsQA and TriviaQA, where it causes more drops on EM and F1 metrics compared to baselines. It means the combination of PAS and DAS is more efficient than solely editing tokens or adding distracting text.
Noticeably, all methods have fair attack effect for datasets with longer contexts, e.g., NewsQA and TriviaQA, because limited numbers of token-level perturbations or adding a single sentence causes fewer impacts on long texts.
Besides, SpanBERT is more robust with slight accuracy declines due to its larger scale and superior pre-training strategy.



In terms of textual quality, \ourmodel achieves the overall lowest \textbf{PPL} and comparable low values on \textbf{GErr}.
TextFooler usually has the lowest GErr values, as it makes pure token-level perturbation that generates fewer sentence-level unnatural errors.
While T3 always generates sentence-level distractors that are meaningless without a complete syntactic structure, resulting in worse performance on GErr and PPL. TASA fulfills trade-off attacks on both token and sentence levels, avoiding significant textual quality loss.

It is also worth mentioning that TASA is better than AddSent at fooling models. Despite having a better textual quality by adding human-annotated distracting texts, samples in AddSent does not perturb the influential part of the original context, limiting its effects on making attacks.


\medskip

\noindent \textbf{Human evaluation.} We randomly sample 150 sets of adversarial samples, each containing 3 samples generated by TextFooler, T3, and \ourmodel originated respectively from the same sample in SQuAD 1.1, using BERT as the victim model. Each set is evaluated in two aspects: (1) Answer preservation, whether the gold answer of a sample remains unchanged; (2) Textual quality, ranking the quality (1 $\sim$ 3) of the context based on the fluency and grammaticality. Totally 63 non-expert annotators are involved, and related results are summarized in Table~\ref{tab:human}. Although \ourmodel is weaker than T3 in answer preservation as T3 always retains the original part of the context, it is equivalent to Textfooler and both of them have a significantly better textual quality than T3 due to the reason we have concluded before. Such a comparable sample quality is sufficient to verify the superiority of \ourmodel, considering its much stronger capability to deceive models (Refer to Appendix~\ref{app:cases} for qualitative adversarial samples by \ourmodel).

\begin{table}[t!]
\setlength{\tabcolsep}{1.3mm}
\setlength{\abovecaptionskip}{0.2cm}
\setlength{\belowcaptionskip}{-0.2cm}
\small
\centering
\begin{tabular}{lccc}
\toprule
Methods & TextFooler & T3 & TASA \\
\midrule
Answer preservation & 79.9$\pm$4.5 & 85.9$\pm$3.3 & 79.1$\pm$4.7 \\
Avg. quality rank & 1.52$\pm$0.06 & 2.64$\pm$0.07 & 1.83$\pm$0.06 \\
\bottomrule
\end{tabular}
\caption{\label{tab:human}Human evaluation results on SQuAD 1.1 (answer preservations are in percentage). $\pm$ indicates the confidence intervals with a 95\% confidence level. }
\end{table}

\subsection{Ablation Studies}
\label{sec:exp_ablation}

We verify the effectiveness of each key module in \ourmodel by: 1) \textit{w/o remove coref.}: without removing coreferences; 2) \textit{w/o PAS}: without applying perturbed answer sentence; 3) \textit{w/o DAS}: without adding distracting answer sentence. The upper part of Table~\ref{tab:ablation} proves their contributions. It can be found that \textit{remove coref.} slightly benefits the quantity of suitable attack samples, while both PAS and DAS make vital contributions to successful attacks and feasible numbers of adversarial samples. 

\begin{figure*}[!t]
    \setlength{\abovecaptionskip}{0.1cm}
    \setlength{\belowcaptionskip}{-0.2cm}
    \centering
    \includegraphics[width=0.8\linewidth]{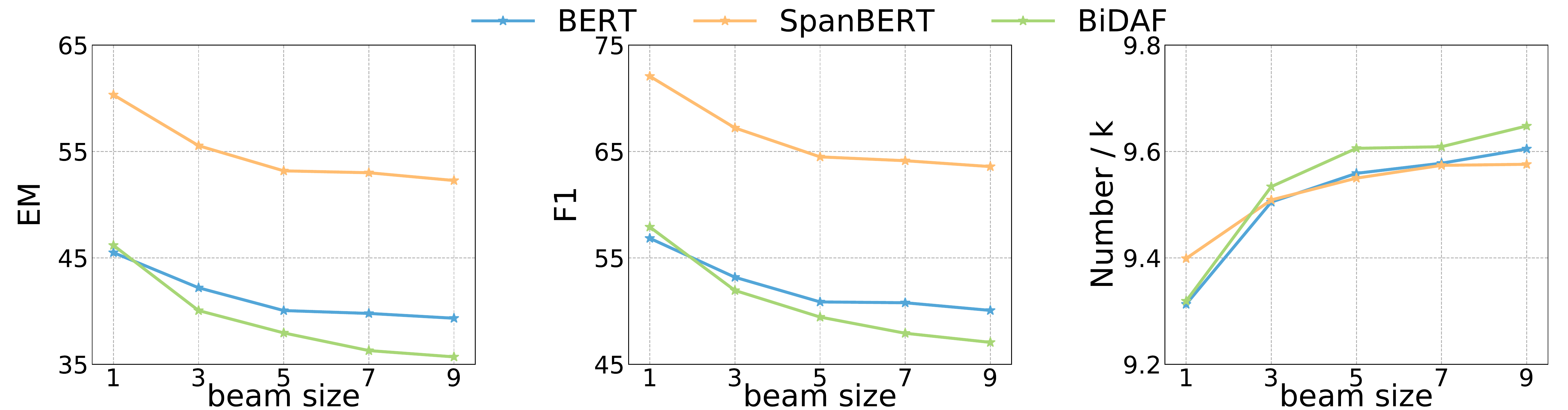}
    \caption{The EM, F1 and quantities of adversarial samples using different beam size on three victim models.}
    \label{fig:beam_size}
\end{figure*}

\begin{table}[t!]
\setlength{\tabcolsep}{1.3mm}
\setlength{\abovecaptionskip}{0.1cm}
\setlength{\belowcaptionskip}{-0.2cm}
\small
\centering
\begin{tabular}{lccccc}
\toprule
\textbf{Modules} & \textbf{EM$\downarrow$} & \textbf{F1$\downarrow$} & \textbf{GErr$\downarrow$} & \textbf{PPL$\downarrow$} & \textbf{Num} \\
\midrule
\ourmodel & 40.06 & 50.87 & 2.98 & 41.15 & 9,559 \\
\midrule
\textit{w/o remove coref.} & 39.95 & 50.39 & 2.96 & 41.13 & 9,374 \\
\textit{w/o PAS} & 59.63 & 70.91 & 2.73 & 35.89 & 8,709 \\
\textit{w/o DAS} & 54.13 & 67.68 & 3.03 & 53.39 & 5,646\\
\midrule
\textit{w/o importance} & 41.44 & 52.32 & 3.01 & 41.94 & 9,564 \\
\textit{w/o quality} & 38.70 & 49.18 & 3.36 & 44.46 & 9,654 \\
\textit{Only use WordNet} & 43.19 & 54.12 & 3.00 & 41.15 & 9,262 \\
\textit{Only use PPDB} & 45.08 & 56.35 & 2.91 & 37.19 & 9,482 \\
\midrule
\textit{w/o edit answer} & 57.63 & 68.91 & 2.86 & 37.00 & 9,559 \\
\textit{Only NEs} & 40.79 & 51.88 & 3.10 & 42.95 & 8,822 \\
\textit{Only nouns} & 43.95 & 55.45 & 3.34 & 45.46 & 7,426 \\
\bottomrule
\end{tabular}
\caption{\label{tab:ablation}Results of \ourmodel ablation studies on SQuAD 1.1 dataset using BERT as the victim model.}
\end{table}

We then do ablations on PAS, including: 1) \textit{w/o importance}: without ranking keywords and edit them randomly; 2) \textit{w/o quality}: without filtering perturbed texts using quality index $U_n$; 
3) \textit{Only use WordNet} as the synonym source; and 4) \textit{Only use PPDB} as the synonym source. 
Based on the middle part of Table~\ref{tab:ablation}, \textit{w/o importance} slightly lower the overall performance. Despite \textit{w/o} quality can promote the attack success rate, it introduces extra textual quality degeneration. Besides, more synonym sources mean a larger search space, so we introduce both WordNet and PPDB into \ourmodel.

Ablations on DAS are finally conducted, 1) \textit{w/o pseudo answer}: do not change answers in DASs; 2) \textit{Only NE} and 3) \textit{Only nouns}: only edit named entities/nouns. Related results are given in the lower Table~\ref{tab:ablation}. The obvious change on \textit{w/o pseudo answer} illustrates that changing the original answer in DASs is crucial for attacking, also proving DAS can shift models' focus from the original answer sentence as they can still derive the gold answer from DASs. Moreover, involving various editing types, including both NE and nouns, benefit the attack effectiveness and generated sample quantity.

\subsection{More Analysis}\label{sec:more_analysis}

\noindent \textbf{Effect of beam size.} 
We vary the beam size during generating PASs and DASs to investigate its influence.
Figure~\ref{fig:beam_size} reports the cahnges of EM, F1, and quantities of adversarial samples. Clearly, a larger beam size leads to better performance and more diverse adversarial samples. Naturally, the larger the beam size also means the slower speed.
Thus, we use $M=5$ for a trading off of performance and efficiency, as we see limited performance gains from beam sizes larger than 5.

\medskip

\noindent \textbf{Shift to the pseudo answers.} Since DAS aims to misguide the attention of models from the original answer sentences to them, we expect QA models to output the pseudo answers contained in DASs. Table~\ref{tab:pseudo_answer} shows the F1 scores between the predicted answers and the pseudo answers on all adversarial samples that include DAS from 5 datasets. The results demonstrate that there are high overlaps between incorrect predictions by victim models and pseudo answers, as these values are close to the performance drops caused by adversarial samples, confirming that DASs can draw attention from models to make incorrect predictions.

\begin{table}[t!]
\setlength{\tabcolsep}{1mm}
\setlength{\abovecaptionskip}{0.1cm}
\setlength{\belowcaptionskip}{-0.2cm}
\small
\centering
\begin{tabular}{lccccc}
\toprule
Datasets & SQuAD 1.1 & NewsQA & NQ & Hotpot & Trivia \\
\midrule
BERT & 39.19 & 20.95 & 36.22 & 36.15 & 24.78 \\
SpanBERT & 33.20 & 20.92 & 32.14 & 38.09 & 27.71 \\
\bottomrule
\end{tabular}
\caption{\label{tab:pseudo_answer}F1 score of predicted answers and pseudo answers, on adversarial samples from \ourmodel with DASs.}
\end{table}

\begin{figure*}[!t]
    \setlength{\abovecaptionskip}{0.1cm}
    \setlength{\belowcaptionskip}{-0.2cm}
    \centering
    \includegraphics[width=0.8\linewidth]{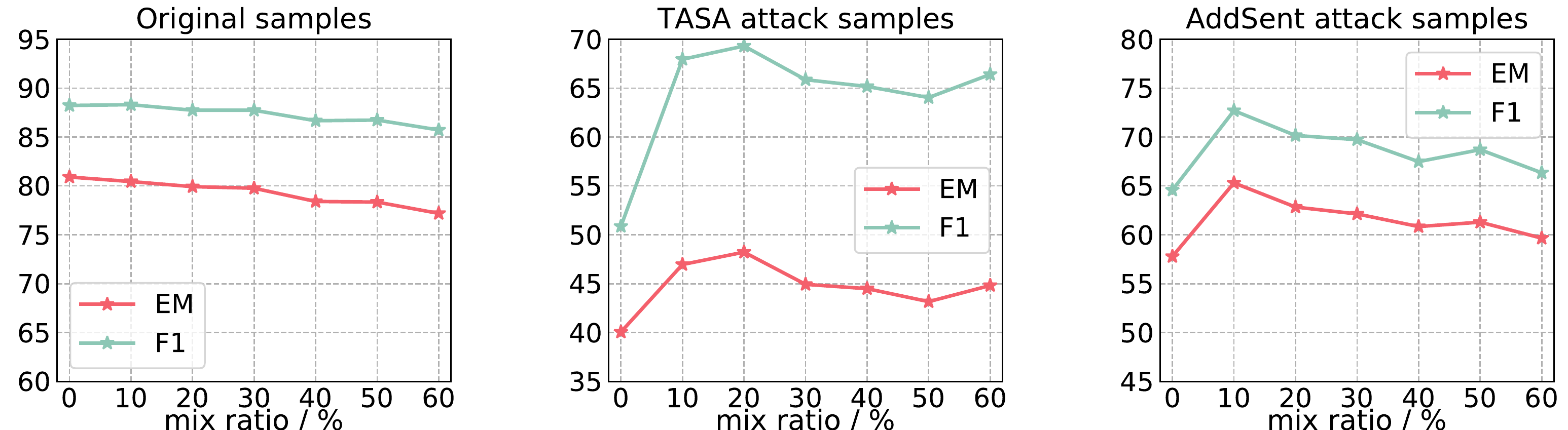}
    \caption{The performance of BERT model fine-tuned on the original SQuAD data \textbf{mixed with adversarial samples from TASA} in different ratios, evaluated on the original dev samples, adversarial samples from TASA and AddSent. We expect a slight influence on original ones, while promotions on the latter two kinds of samples.}
    \label{fig:adv_train}
\end{figure*}

\medskip

\noindent \textbf{Adversarial training.} 
To verify the effectiveness of TASA in improving the robustness of QA models, we randomly replace training data in SQuAD~1.1 with corresponding adversarial samples generated by TASA in varied ratios, and then use the new training data to fine-tune a BERT model. The performance on the original dev set, the adversarial dev set generated by TASA, and samples from AddSent, is shown in Figure~\ref{fig:adv_train}, where different mixing ratios are used, 
Noticeably, with a suitable mixture ratio, adversarial samples from TASA can make models more robust under adversarial attacks without significant performance loss on the original data. Interestingly, this defense capability can also be transferable to other adversarial data, e.g. AddSent. Such results verify the potential of \ourmodel to enhance the current QA models.

\section{Related Work}

\noindent
\textbf{Question answering.}
Extractive QA is the most common QA task, where the answer is a text span in the supporting context. There are various datasets, e.g., SQuAD, NewsQA, and NaturalQuestions~\cite{squad1.1,squad2.0,newsqa,natural_questions}, motivating more works on QA models, such as end2end models like BiDAF, R-Net, QANet and so on~\cite{bidaf,rnet,qanet}. Pre-trained models are widely applied recently, such as BERT, RoBerta, and SpanBert~\cite{bert,roberta,spanbert}. They realize remarkable promotions benefited from huge corpora, meanwhile they can also be used as backbones to solve more complex QA tasks~\cite{cao2019bag,huang2021dagn}. Nevertheless, there are more concerns~\cite{unnatural_nli,what_bert_learn,universal_trigger} whether models can really capture contextual information rather than using token-level knowledge simply.

\medskip

\noindent
\textbf{Textual adversarial attack.}
Textual adversarial attack has been widely investigated in general tasks like text classification and natural language inference (NLI). Some works use character-level misspelled tokens to attack models, but are easy to be defended~\cite{character_attack,hotflip,textbugger,defend_misspell}.
More studies use more sophisticated toke-level perturbations~\cite{pwws,generate_adversarial,word_level_attack,clare} or phrase/sentence-level editing~\cite{adversarial_syntatic,maya,lei2022phrase} to produce adversarial texts, with some filtering strategies to guarantee the text meaning and quality. However, none of them shows their effectiveness on QA tasks.

There are some efforts on attacking QA models. AddSent~\cite{addsent} contains adversarial samples with distracting sentences added by human annotators. \citeauthor{human_in_loop} employ human testers to interact with models and realize dynamic attacks. Despite showing their effectiveness, these approaches are not extensible and limited in scale. There are also automatic methods. T3 ~\cite{t3} utilizes a Tree LSTM to obtain a distracting sentence based on the skeleton of the question. Universal Trigger~\cite{universal_trigger} find input-agnostic texts that deceive models for a specific question type via gradient-guided search. Our \ourmodel differs from them as it bridges contexts and questions to attack more efficiently and suits more general conditions.

\section{Conclusion}

We present \ourmodel, an automatic adversarial attack method for QA models. It generates twin answer sentences, perturbed answer sentence (PAS), and distracting answer sentence (DAS), to construct a new adversarial context in a QA sample. It can deceive models and misguide them to an incorrect answer based on their pitfalls that overly rely on matching sensitive keywords during predicting answers. 
In experiments, \ourmodel achieves remarkable attack performance on five datasets and three victim models with satisfactory sample quality. Our additional analysis also proves that it is possible to get more robust QA models via \ourmodel in the future.

\section*{Acknowledgements}
This work is supported in part by Australian Research Council FL-170100117, IC-190100031, and LE-200100049.

\bibliography{anthology,custom}
\bibliographystyle{acl_natbib}

\newpage
\appendix

\section{Implementation Details}

\subsection{Training Victim Models}

\noindent \textbf{BERT} We use the huggingface-transformers\footnote{https://github.com/huggingface/transformers} to implement the model and the \textit{bert-base-uncased} version of BERT model\footnote{https://huggingface.co/bert-base-uncased} to initialize the model weights. It contains 12 layers with a hidden size of 768. A linear layer is added to predict the start and end positions of the answer span. 

During fine-tuning BERT on different QA datasets, we set the maximum input sequence length as 384, using an Adam optimizer whose initial learning rate is 6.25e$-5$ with the batch size 32. The epoch number is 3 and the final model after all epochs will be saved as the victim model.

\noindent \textbf{SpanBERT} We also use the huggingfance-transformers to implement the model, along with \textit{spanbert-large-cased} version\footnote{https://huggingface.co/SpanBERT/spanbert-large-cased} to initialize the weights. It contains 24 layers with a hidden size of 1024. A linear layer is added to predict the start and end positions of the answer span. 

During finetuning, we set the maximum input sequence length as 512, using an Adam optimizer whose initial learning is rate 2e$-5$ with the batch size 32. The epoch number is 3 and the final model after all epochs will be saved as the victim model.

\noindent \textbf{BiDAF} We use the model implementation provided by AllenNLP\footnote{https://github.com/allenai/allennlp}. The 6B 100d version of GLoVe~\cite{glove} is used to initialize the token embedding layer of BiDAF.

During training, we set the maximum input context length as 800, using an Adam optimizer with an initial learning rate 1e-3 and batch size 40 to train BiDAF for 20 epochs. All other settings are in default. We will save the model with the best performance on the dev set as the victim model.

\subsection{\label{app:implementation}TASA}

\noindent \textbf{Remove coreferences.} We use NeuralCoref\footnote{https://github.com/huggingface/neuralcoref} combined with SpaCy\footnote{https://spacy.io} to find out the coreferences in contexts.

\noindent \textbf{Perturbation on answer sentences.} To select the answer sentence $s_a$, we use the answer span position given by each label in datasets, where the sentence containing this span is regarded as $s_a$. If the answer spans are not unique, we use the answer span that is chosen most times by annotators or the first span in the context.
The lemmas and POS tags of different are obtained via SpaCy. The POS tag set used to get keywords is \{\textit{"VERB", "NOUN", "ADJ", "ADV"}\}. When perturbing a token with its synonyms, we use pyinflect\footnote{https://spacy.io/universe/project/pyinflect/} to recover the lemmas of replacements into the same inflections of the original token.

\noindent \textbf{Adding distracting answer sentences.} We construct a NER dictionary and a word dictionary (except named entities) for each target dataset by parsing all contexts in both the train and dev sets via SpaCy. During generating DAS or changing answers in DAS, we randomly sample named entities with the same NER tag or words with the same POS tag from the dictionaries we built before. Each time, we sample $N=20$ from them and ensure there is no overlap with the original entity/token we want to replace. Pyinflect is also used during replacement.

\noindent \textbf{Beam Search.} During beam search, we apply an early-stop strategy on the filtered results after each time of a search. We also restrict the maximum perturbation number to 5 for both PAS and DAS. If one of the following 3 criteria is satisfied: 1) the minimum effect score $E_n$ among them satisfies $min(E_n) \geq T_E$, where $T_E$ is a threshold and $T_E=0.2$; 2) all possible token/entities have been replaced, the beam search will stop, and the final $M$ sentences will proceed to the next step.

\noindent \textbf{Quality filtering.} During filtering, we use the official USE model\footnote{https://tfhub.dev/google/universal-sentence-encoder/4} to get USE similarity and a small size GPT2 model\footnote{https://huggingface.co/gpt2} to get the PPL. 

\noindent \textbf{Model to determine whether a question is answerable for modified context.} We use a RoBerta$_{\text{base}}$ model fine-tuned on the original SQuAD 2.0 dataset\footnote{https://huggingface.co/deepset/roberta-base-squad2} as the answerable judgment model for SQuAD 1.1, because these two datasets share the same corpus and model trained on SQuAD 2.0 has the capability to predict whether a question is answerable. If the model outputs the highest answer possibility on the special ``<s>'' token at the beginning of the input, then the current sample is regarded as unanswerable. 

For the rest four datasets, we use other Roberta models fine-tuned on our newly constructed training sets. More specially, each set includes the original samples whose label is answerable and negative samples (unanswerable samples) whose quantity is the same as answerable samples.  Here, each negative sample has a question obtained by randomly sampling from the whole dataset that does not belong to the given context.
We follow the same training pattern as SQuAD 2.0 to fine-tune RoBerta models, where the model needs to have the capability of both answering answerable samples and output \textit{"unanswerable"} label for unanswerable samples. We list the performance of all these models used in our experiments in Table~\ref{tab:determine_model_performance}. Our constructed data are less challenging for models because the questions of negative samples are randomly sampled from the whole corpus, which may be quite different from the context and easy to be distinguished.

\begin{table}[t!]
\small
\setlength{\abovecaptionskip}{0.1cm}
\setlength{\belowcaptionskip}{-0.3cm}
\setlength{\tabcolsep}{4pt}
\centering
\begin{tabular}{lrrrrr}
\toprule
Datasets & Has EM & Has F1 & No EM & EM & F1 \\
\midrule
SQuAD 2.0* & 77.94 & 84.03 & 81.80 & 79.87 & 82.91 \\
NewsQA & 52.68 & 65.64 & 98.43 & 75.56 & 82.04 \\
NQ & 67.72 & 79.26 & 99.21 & 83.46 & 89.24 \\
HotpotQA & 59.43 & 77.31 & 99.83 & 79.63 & 88.57 \\
TriviaQA & 48.45 & 53.10 & 99.88 & 74.17 & 76.49\\
\bottomrule
\end{tabular}
\caption{\label{tab:determine_model_performance}The performance of our RoBerta models to determine whether a question is answerable on our newly constructed dataset containing answerable and unanswerable samples. ``Has'' means samples has an answer, ``No'' means samples without an answer. *: we directly use the SQuAD 2,0 dataset to train the model for SQuAD 1.1 as they share the same corpus.}
\end{table}

\begin{table}[t!]
\setlength{\tabcolsep}{1.5mm}
\setlength{\abovecaptionskip}{0.1cm}
\setlength{\belowcaptionskip}{-0.3cm}
\centering
\begin{tabular}{lr}
\toprule
Hyperparameters & Value \\
\midrule
effect score threshold $T_E$ & 0.2 \\
quality score threshold $T_U$ & -2 \\
beam search size $M$ & 5 \\
random sampling size for DAS $N$ & 20\\
\bottomrule
\end{tabular}
\caption{\label{tab:hyperparameters}Values of hyperparameters used in TASA.}
\end{table}

We list all hyperparameter values used by TASA method in Table~\ref{tab:hyperparameters}, which are obtained by empirical tuning based on the trade-off between attack effectiveness and textual quality.
We conduct all our experiments on a single NVIDIA V100 GPU. We also publish our code anonymously at \url{https://anonymous.4open.science/r/TASA/}.

\textbf{The possible limitations of our method}: \ourmodel is only appliable to extractive QA tasks, and the question or answer is not perturbed to achieve a better deception on models ,which we leave for the future work. 

\subsection{Baselines}

We run the official code provided by the authors of original baseline papers to derive the relevant results in our experiments. We have tried our best to reproduce the results reported in papers, but their configurations are quite different from ours.

\noindent \textbf{TextFooler} Since this method is not designed for QA tasks, we made some modifications to it. 1) We only use the context as the targeted attack text and mask tokens within it to get their importance scores; 2) in order to avoid changing the answer, we do not involve answer tokens as the perturbation targets; 3) we also use the prediction possibility on the gold answer to get the evaluation on each time attack and determine when to stop the attack. We implement our attack based on the official code and keep other settings as the default.

\noindent \textbf{T3} We apply its official code directly as it already contains the function to attack QA dataset in SQuAD format. To make a fair comparison, we use its \textbf{black-box} configuration without accessing the internal parameters of models. Besides, we use its \textbf{target} configuration, which aims to specially misguide the model predictions to the pseudo answer in the distracting sentence and shows a better performance.

\subsection{\label{app:dataset}Datasets}

We provide some statistics about 5 datasets we used in Table~\ref{tab:dataset_statistics}. We use the official release version of SQuAD 1.1, while the MRQA version~\footnote{https://github.com/mrqa/MRQA-Shared-Task-2019} for other 4 datasets, where we transform them into the same format as SQuAD 1.1 for the convenience of our experiments.

\begin{table}[t!]
\small
\setlength{\abovecaptionskip}{0.1cm}
\setlength{\belowcaptionskip}{-0.3cm}
\setlength{\tabcolsep}{4pt}
\centering
\begin{tabular}{lcccc}
\toprule
Datasets & |Q| & |C| & Train size & Dev size  \\
\midrule
SQuAD 1.1 & 11 & 137 & 87,599 & 10,570 \\
NewsQA & 8 & 599 & 74,160 & 4,212 \\
Natural Questions (NQ) & 9 & 153 & 104,071 & 12,836\\
HotpotQA & 22 & 232 & 72,928 & 5,901 \\
TriviaQA & 16 & 784 & 61,688 & 7,785\\
\bottomrule
\end{tabular}
\caption{\label{tab:dataset_statistics}The statistics of 5 datasets used in our experiments. |C| is the average length of context, |Q| is the average length of questions, both in token level.}
\end{table}

\section{Additional Results}

\begin{table}[t!]
\setlength{\abovecaptionskip}{0.1cm}
\setlength{\belowcaptionskip}{-0.3cm}
\setlength{\tabcolsep}{2.8pt}
\small
\centering
\begin{tabular}{ll|rrrrr}
\toprule
\textbf{Dataset} & \textbf{method} & \textbf{EM$\downarrow$} & \textbf{F1$\downarrow$} & \textbf{GErr$\downarrow$} & \textbf{PPL$\downarrow$} & \textbf{Num}\\
\midrule
\multirow{4}{0pt}{SQuAD 1.1} & Original & 65.72 & 75.97 & 2.39 & 33.25 & 10,570 \\
& AddSent* & 40.87 & 49.19 & 2.47 & 33.98 & 3,560 \\
& TextFooler & 42.65 & 56.96 & \textbf{2.56} & \textbf{37.95} & 7,228 \\
& T3 & 52.74 & 61.69 & 4.44 & 44.20 & 9,681 \\
& OURS & \textbf{37.96} & \textbf{49.44} & 2.89 & 41.08 & 9,606 \\
\midrule
\multirow{4}{0pt}{News QA} & Original & 43.99 & 57.64 & 1.98 & 22.50 & 4,212 \\
& TextFooler & \textbf{32.03} & \textbf{46.69} & \textbf{2.11} & 23.92 & 3,662 \\
& T3 & 39.21 & 51.89 & 2.56 & 22.99 & 3,775 \\
& OURS & 33.76 & 47.23 & 2.19 & \textbf{22.83} & 2,903 \\
\midrule
\multirow{4}{0pt}{NQ} & Original & 56.77 & 68.83 & 20.48 & 49.74 & 12,836 \\
& TextFooler & 39.65 & 53.91 & \textbf{20.50} & 47.31 & 7,111 \\
& T3 & 41.98 & 52.27 & 20.72 & 65.61 & 10,460 \\
& OURS & \textbf{37.86} & \textbf{49.56} & 20.58 & \textbf{43.25} & 8,955 \\
\midrule
\multirow{4}{0pt}{Hotpot QA} & Original & 46.38 & 63.88 & 3.73 & 17.01 & 5,901 \\
& TextFooler & 36.75 & 55.40 & \textbf{3.87} & 18.49 & 4,974 \\
& T3 & 41.38 & 58.41 & 5.16 & 20.78 & 5,186 \\
& OURS & \textbf{34.12} & \textbf{49.15} & 4.00 & \textbf{15.31} & 5,050 \\
\midrule
\multirow{4}{0pt}{Trivia QA} & Original & 45.19 & 52.85 & 3.74 & 24.42 & 7,785 \\
& TextFooler & \textbf{38.05} & \textbf{45.25} & \textbf{3.82} & 25.63 & 7,227 \\
& T3 & 43.22 & 50.07 & 4.20 & 25.50 & 7,434 \\
& OURS & 40.21 & 46.47 & 3.87 & \textbf{24.72} & 7,110 \\
\bottomrule
\end{tabular}
\caption{\label{tab:bidaf_results}Attack results on 5 QA datasets using BiDAF as the victim model. The best results are \textbf{bold}. \textbf{Num} is the sample number of a dataset or generated from the whole dataset by a method. $\downarrow$ represents that the lower the better. *: annotated by humans. }
\end{table}

\subsection{Using BiDAF as the Victim Model}
\label{app:bidaf_results}

We also include BiDAF as one kind of victim model in our experiments, as it is a representative End2end RNN-based model. The related results are not provided in the main part due to the page limitation and its current fair performance compared to SOTA models. The attack results on five dataset same as \secref{sec:exp} are shown in Table~\ref{tab:bidaf_results}. Similarly, our \ourmodel achieves the best attack effectiveness in 3 datasets among 5 according to the declining scale of EM and F1, while remaining comparable in the other 2 datasets. In addition, \ourmodel also achieves the overall lower PPL among all conditions and a close performance to TextFooler in terms of grammar error. These observations again demonstrate the superiority of our method. Moreover, it is noticeable that BiDAF is less vulnerable than BERT as the performance degeneration is slighter, especially on datasets with long contexts, e.g., NewsQA and TriviaQA.

\subsection{Shift to Pseudo Answers}

Since PASs aim to attract models' focus from original answer sentences and misguide models to make predictions on pseudo answers. We have conducted related experiments in \secref{sec:more_analysis} to prove their validity. Here, we provide more results about this experiment, including not only the F1 scores between predictions by 3 models on TASA adversarial samples and the pseudo answers contained in the corresponding PASs, but also F1 scores between pseudo answers and the models' predictions on the original samples, making a further comparison to eliminate the possible influence of the existing prediction overlap. Results are given in Table~\ref{tab:additional_pseudo_answer}. Obviously, all models under all conditions tend to predict answers that have more overlaps with pseudo answers given TASA adversarial samples, proving the misguiding effect of DASs. Besides, the F1 score difference between predictions on TASA samples and the original samples will be reduced on datasets where the attack capability of TASA is consistently weaker, such as NewsQA and TriviaQA. This proves that the efficiency of DASs drawing models' attention affects the attack performance remarkably when combined with PAS.

\begin{table}[t!]
\setlength{\tabcolsep}{1mm}
\setlength{\abovecaptionskip}{0.1cm}
\setlength{\belowcaptionskip}{-0.3cm}
\setlength{\tabcolsep}{1.5pt}
\small
\centering
\begin{tabular}{lccccc}
\toprule
Datasets & SQuAD~1.1 & NewsQA & NQ & Hotpot & Trivia \\
\midrule
BERT & 39.19 & 20.95 & 36.22 & 36.15 & 24.78 \\
BERT(Ori) & 15.08 & 14.40 & 21.21 & 20.38 & 20.30 \\
\midrule
SpanBERT & 33.20 & 20.92 & 32.14 & 38.09 & 27.71 \\
SpanBERT(Ori) & 16.06 & 16.54 & 22.49 & 21.71 & 23.54 \\
\midrule
BiDAF & 26.34 & 16.69 & 29.43 & 27.80 & 20.08 \\
BiDAF(Ori) & 12.68 & 12.80 & 18.51 & 16.76 & 16.42 \\
\bottomrule
\end{tabular}
\caption{\label{tab:additional_pseudo_answer}F1 score between predicted answers on TASA adversarial samples with DASs and pseudo answers from corresponding DASs, or between predicted answers on the original samples and pseudo answers (Ori), using 3 victim models on 5 datasets.}
\end{table}

\begin{figure}[!t]
    \setlength{\abovecaptionskip}{0.1cm}
    \setlength{\belowcaptionskip}{-0.3cm}
    \centering
    \includegraphics[width=1\linewidth]{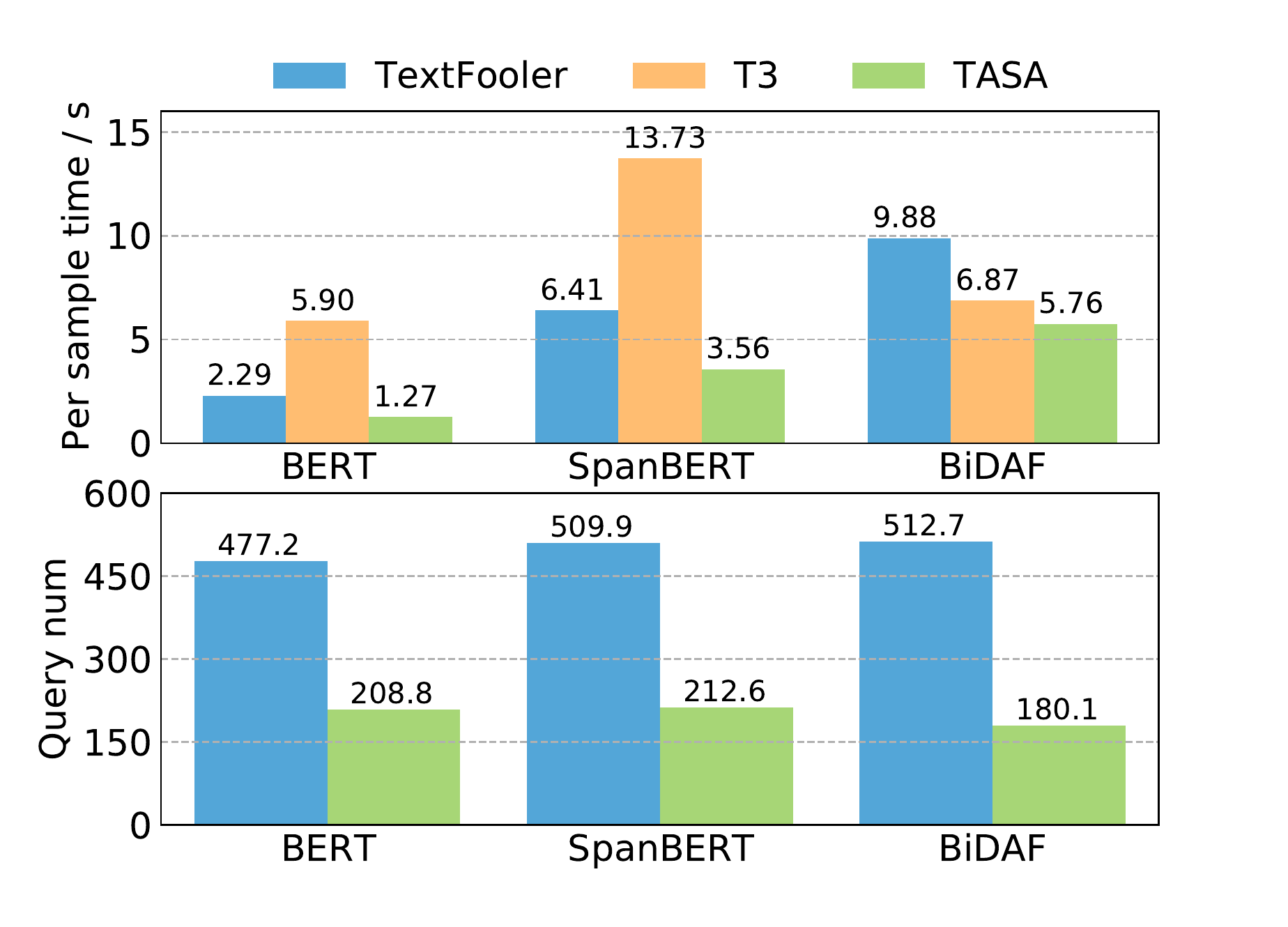}
    \caption{The per sample time to generate adversarial samples (in second) and average query number to victim models of TextFooler, T3 and \ourmodel, using all kinds of victim models on SQuAD 1.1 dataset.}
    \label{fig:complexity}
\end{figure}

\begin{table*}[t!]
\setlength{\abovecaptionskip}{0.1cm}
\setlength{\belowcaptionskip}{-0.3cm}
\setlength{\tabcolsep}{1.5mm}
\small
\centering
\begin{tabular}{ll|rrr|rrr|rrr}
\toprule
\textbf{Dataset} & \textbf{source} & \multicolumn{3}{c|}{BERT} & \multicolumn{3}{c|}{SpanBERT} & \multicolumn{3}{c}{BiDAF} \\
\cmidrule{3-11}
 & & ratio~/~\% & EM & F1 & ratio~/`\% & EM & F1 & ratio~/~\% & EM & F1 \\
\midrule
\multirow{3}{*}{SQuAD 1.1} & PAS+DAS & 50.2 & 22.75 & 32.98 & 53.2 & 41.67 & 53.31 & 54.4 & 26.47 & 36.46 \\
& PAS & 8.9 & 51.06 & 63.48 & 10.2 & 63.90 & 78.29 & 9.3 & 37.46 & 50.08 \\
& DAS & 40.9 & 58.86 & 70.03 & 36.6 & 70.36 & 81.12 & 36.3 & 55.28 & 68.69 \\
\midrule
\multirow{3}{*}{NewsQA} & PAS+DAS & 40.8 & 32.42 & 45.77 & 42.7 & 43.47 & 59.25 & 44.0 & 25.84 & 39.38 \\
& PAS & 19.9 & 34.23 & 47.80 & 20.7 & 44.67 & 60.67 & 21.1 & 26.49 & 40.89 \\
& DAS & 39.3 & 50.01 & 64.52 & 36.6 & 62.85 & 75.29 & 34.9 & 40.95 & 53.78 \\
\midrule
\multirow{3}{*}{NQ} & PAS+DAS & 54.1 & 33.09 & 45.27 & 54.5 & 44.12 & 58.50 & 54.9 & 30.29 & 42.25 \\
& PAS & 19.5 & 47.55 & 62.97 & 19.9 & 53.21 & 69.24 & 19.9 & 41.24 & 55.81\\
& DAS & 26.4 & 60.80 & 70.25 & 25.6 & 64.26 & 74.94 & 25.2 & 51.72 & 60.59 \\
\midrule
\multirow{3}{*}{HotpotQA} & PAS+DAS & 61.1 & 21.60 & 32.63 & 63.5 & 38.55 & 54.86 & 60.8 & 29.91 & 44.85 \\
& PAS & 14.8 & 33.10 & 46.91 & 14.9 & 51.52 & 68.81 & 14.9 & 34.65 & 50.73 \\
& DAS & 24.1 & 39.53 & 52.78 & 21.6 & 56.55 & 72.53 & 24.3 & 44.34 & 59.02 \\
\midrule
\multirow{3}{*}{TriviaQA} & PAS+DAS & 62.9 & 48.07 & 51.73 & 63.3 & 60.68 & 66.59 & 63.5 & 37.10 & 43.13  \\
& PAS & 13.2 & 51.86 & 54.42 & 13.3 & 63.16 & 69.37 & 12.3 & 40.61 & 47.29 \\
& DAS & 23.9 & 60.01 & 64.72 & 23.4 & 71.41 & 77.31 & 24.2 & 48.70 & 55.65 \\
\bottomrule
\end{tabular}
\caption{\label{tab:composition_performance} The ratio and performance of QA models on different compositions of adversarial samples generated by TASA, on all 5 datasets and 3 victim models. PAS+DAS: both PAS and DAS are applicable in the current sample; PAS: only PAS is applicable in the current sample; DAS: only DAS is applicable in the current sample.}
\end{table*}

\subsection{Analysis of Computational Complexity}

We illustrate the per sample attack time and query number to the victim models of our \ourmodel and two baselines, TextFooler and T3, on SQuAD 1.1 dataset and all 3 types of models, in Figure~\ref{fig:complexity}. Note that T3 has a constant query number to victim models, so it is not involved in this part. All results are obtained on a single NVIDIA V100 GPU.
It can be seen that our \ourmodel is the fastest attack method compared to other baselines, and it also makes fewer queries to the victim model before obtaining an adversarial sample. Although T3 has a constant query number to the victim model, its complexity depends on the scale of the target model's embedding.

\subsection{The Composition of Samples Generated by TASA}


Although we design twin sentences, PAS and DAS, to attack QA models, it is possible that not both of them are applicable for a sample. E.g., only PAS is applicable if there is no proper named entity or noun that can be edited in the answer sentence excluding keywords and the gold answer; or only DAS is applicable for a sample where no overlapped keyword is found between the answer sentence and question. A sample where only PAS or DAS is applied will also be put into the final adversarial sample set, along with samples that both PAS and DAS (PAS+DAS) are involved. 
In order to study the compositions of different adversarial sample sources, as well as the performance of victim models on each part, 
we provide the ratios of each type of samples generated by TASA on different datasets along with the performance of QA models on them in Table~\ref{tab:composition_performance}. 

It can be found that PAS+DAS compose the majority of adversarial samples on nearly all datasets, while the quantities samples that only contain DAS are generally larger than samples with only PAS. 
When comes to the performance of QA models on each part, it can be found that PAS+DAS has the best attack effectiveness among all types of samples, because they not only deceive models using perturbed keywords but also utilize distracting answer sentences to misguide models to make wrong predictions on the included pseudo answers.
On the other hand, only using PAS or DAS can lower the attack effectiveness.
The reason is that a single attack source may not sufficiently fool models, proving the necessity of combining the two folds of pitfall we discussed in \secref{sec:prove_experiment} into the adversarial attack on the QA task.
Moreover, the attack difference between PAS+DAS and PAS will be narrowed on datasets having longer contexts like NewsQA and TriviaQA, where EM and F1 values on these two types of samples are more close. The relatively weak attack ability on such datasets should be the main cause. Besides, longer input sequences will lower the attention weights of models on each token, merely adding PAS also results in less influence because their ratio on the whole input becomes smaller. 

\section{Qualitative Samples}\label{app:cases}

\begin{figure}[h]
    \setlength{\abovecaptionskip}{-0cm}
    \setlength{\belowcaptionskip}{-0.4cm}
    \centering
    \includegraphics[width=1\linewidth]{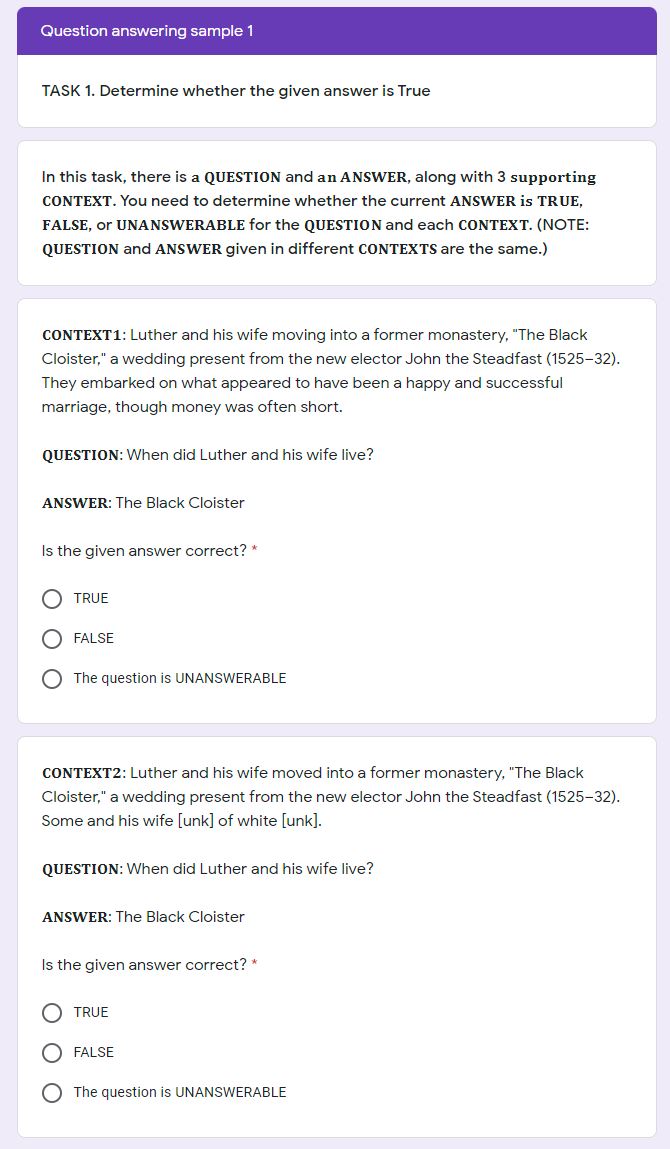}
    \caption{Screenshot of instructions for human evaluation (part1).}
   \label{fig:human_anno1}
\end{figure}

\begin{figure}[!t]
    \setlength{\abovecaptionskip}{-0cm}
    \setlength{\belowcaptionskip}{-0.4cm}
    \centering
    \includegraphics[width=1\linewidth]{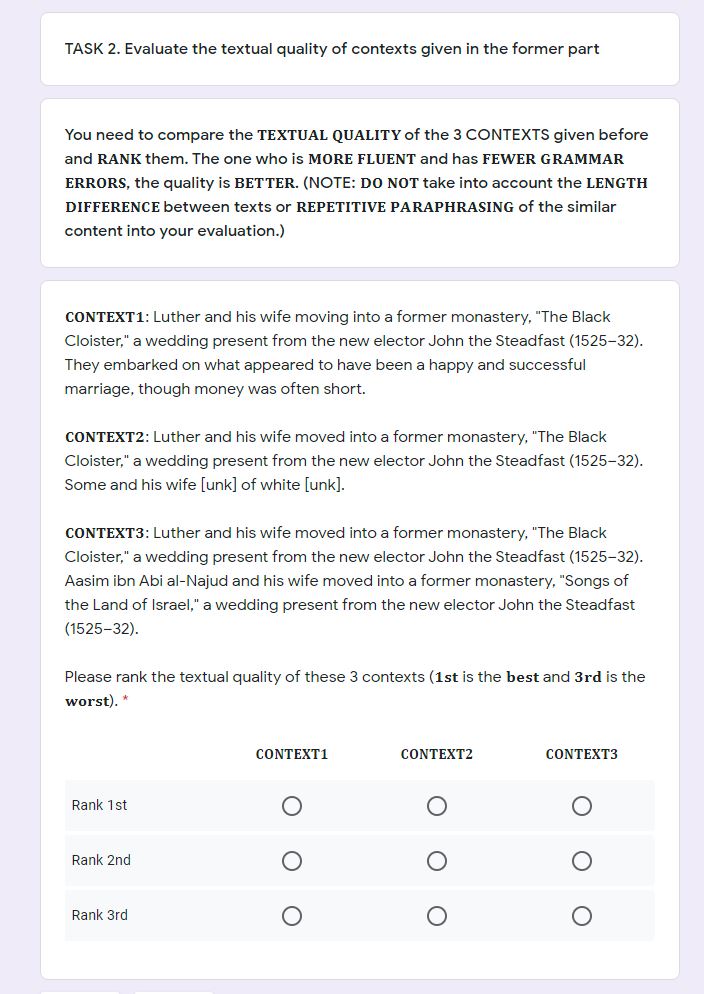}
    \caption{Screenshot of instructions for human evaluation (part2).}
  \label{fig:human_anno2}
\end{figure}

We provide some samples generated by TextFooler, T3 and TASA along with corresponding model predictions in Table~\ref{tab:case1}, Table~\ref{tab:case2}. We also provide the instruction screenshot for human evaluation in Figure~\ref{fig:human_anno1} and Figure~\ref{fig:human_anno2}.

\begin{table*}[t!]
\small
\setlength{\abovecaptionskip}{0.2cm}
\setlength{\belowcaptionskip}{-0.2cm}
\centering
\begin{tabularx}{1\textwidth}{lX}
\toprule
\textbf{Original context} & Long-term active memory is acquired following infection by activation of B and T cells. \underline{Active immunity can also be generated artificially, through \setorange{vaccination}.} The principle behind vaccination (also called immunization) is to introduce an antigen from a pathogen in order to stimulate the immune system and develop specific immunity against that particular pathogen without causing disease associated with that organism. This deliberate induction of an immune response is successful because it exploits the natural specificity of the immune system, as well as its inducibility. With infectious disease remaining one of the leading causes of death in the human population, vaccination represents the most effective manipulation of the immune system mankind has developed.  \\
\textbf{Question} & By what process can active immunity be generated in an artificial manner? \\
\textbf{Answer} & vaccination \\
\hdashline
\textbf{TextFooler context} & Long-term active memory is \setblue{obtaining} following infection by activation of B and T cells. \underline{Active immunity can also \setblue{constitute} generated \setblue{mannually}, through \setorange{vaccination}.} The principle behind vaccination (also called immunization) is to introduce an antigen from a pathogen in order to stimulate the immune system and develop specific immunity against that particular pathogen without causing disease associated with that organism. This deliberate induction of an immune response is successful because it exploits the natural specificity of the immune system, as well as its inducibility. With infectious disease remaining one of the leading causes of death in the human population, vaccination represents the most effective manipulation of the immune system mankind has developed. \\
\textbf{Model prediction} & vaccination \\
\hdashline
\textbf{T3 context} & Long-term active memory is acquired following infection by activation of B and T cells. \underline{Active immunity can also be generated artificially, through \setorange{vaccination}.} The principle behind vaccination (also called immunization) is to introduce an antigen from a pathogen in order to stimulate the immune system and develop specific immunity against that particular pathogen without causing disease associated with that organism. This deliberate induction of an immune response is successful because it exploits the natural specificity of the immune system, as well as its inducibility. With infectious disease remaining one of the leading causes of death in the human population, vaccination represents the most effective manipulation of the immune system mankind has developed. \setgreen{Active immunity generated immunization.}\\
\textbf{Model prediction} & vaccination \\
\hdashline
\textbf{TASA context} & Long-term active memory is acquired following infection by activation of B and T cells. \underline{\setblue{Alive} immunity can also be \setblue{produced} artificially, through \setorange{vaccination}}. The principle behind immunization (also called immunization) is to introduce an antigen from a pathogen in rank to stimulate the immune system and arise precise resistance against that particular pathogen without causing disease associated with that organism. Thpersonify deliberate induction of an immune response personify successful because it utilises the natural specificity of the immune system of rule, as well as its inducibility. With infectious disease remaining one of the leading causes of death in the human population, vaccination represents the most effective manipulation of the immune system mankind has developed. \setgreen{Active irradiation can also be generated artificially, through sword - cut.} \\
\textbf{Model prediction} & \noindent{\setred{sword - cut}} \\
\bottomrule
\end{tabularx}

\begin{tabularx}{\textwidth}{lX}
\toprule
\textbf{Original context} & In 1873, Tesla returned to his birthtown, Smiljan. Shortly after he arrived, Tesla contracted cholera; he was bedridden for nine months and was near death multiple times. \underline{Tesla's father, in a moment} \underline{of despair, promised to send him to the best engineering school if he recovered from the illness (his} \underline{father had originally wanted him to \setorange{enter the priesthood}).}  \\
\textbf{Question} & What did Tesla's father originally want him to do? \\
\textbf{Answer} & enter the priesthood \\
\hdashline
\textbf{TextFooler context} & In 1873, Tesla returns to his birthtown, Smiljan. Shortly after he arrived, Tesla contracted cholera; he was \setblue{crippled} for nine months and was near death multiple times. \underline{Tesla's \setblue{dads}, in a \setblue{tiempo} of \setblue{angst},} \underline{\setblue{pledging} to \setblue{transmits} him to the \setblue{advisable engineers schooling} if he \setblue{recaptured} from the \setblue{malady} (his} \underline{father had originally wanted him to \setorange{enter the priesthood}).} \\
\textbf{Model prediction} & enter the priesthood \\
\hdashline
\textbf{T3 context} & In 1873, Tesla returned to his birthtown, Smiljan. Shortly after he arrived, Tesla contracted cholera; he was bedridden for nine months and was near death multiple times. \underline{Tesla's father, in a moment} \underline{of despair, promised to send him to the best engineering school if he recovered from the illness (his} \underline{father had originally wanted him to \setorange{enter the priesthood}).} \setgreen{Our our father our want father to us entering of ordained.} \\
\textbf{Model prediction} & enter the priesthood \\
\hdashline
\textbf{TASA context} & In 1873, Tesla delivered to his birthtown, Smiljan. Shortly after he arrived, Tesla contracted Asiatic cholera; he was bedridden for nine months and was near death multiple times. \underline{Tesla's \setblue{dad}, in a} \underline{moment of despair, promised to send him to the best engineering school if he recovered from the} \underline{illness (his \setblue{dad} had \setblue{in the beginning required} him to \setorange{enter the priesthood}).} \setgreen{The Bureau of Near Eastern Affairs's father, in a moment of despair, promised to send him to the best engineering school if he recovered from the illness (his father had originally wanted him to sadden the businessman).} \\
\textbf{Model prediction} & \noindent{\setred{sadden the businessman}} \\
\bottomrule
\end{tabularx}
\caption{\label{tab:case1} Adversarial contexts generated by TextFooler, T3, and TASA, compared to the original context on SQuAD 1.1 using BERT as victim model, along with predicted answers by the model. \setorange{Gold answer}, \setblue{perturbed tokens} (i.e. perturbations on answer sentence for TASA),  \setgreen{added distracting sentences} (i.e. DAS for TASA), and \setred{wrong answers} are in different colors. \underline{Underlined sentences} indicate the answer sentences. }
\end{table*}

\begin{table*}[t!]
\small
\setlength{\abovecaptionskip}{0.2cm}
\setlength{\belowcaptionskip}{-0.2cm}
\centering
\begin{tabularx}{\textwidth}{lX}
\toprule
\textbf{Original context} & The Daily Mail newspaper reported in 2012 that the UK government's benefits agency was checking claimants' "Sky TV bills to establish if a woman in receipt of benefits as a single mother is wrongly claiming to be living alone" – as, it claimed, subscription to sports channels would betray a man's presence in the household. \underline{In December, the UK’s parliament heard a claim that a subscription to} \underline{BSkyB was ‘\setorange{often damaging}’ , along with alcohol, tobacco and gambling.} Conservative MP Alec Shelbrooke was proposing the payments of benefits and tax credits on a "Welfare Cash Card", in the style of the Supplemental Nutrition Assistance Program, that could be used to buy only "essentials".  \\
\textbf{Question} & what did the UK parliment hear that a subscription to BSkyB was? \\
\textbf{Answer} & often damaging \\
\hdashline
\textbf{TextFooler context} & The Daily Mail newspapers reported in 2012 that the UK government's benefits agency was checking claimants' "Sky TV bills to establish if a woman in receipt of benefits as a unaccompanied mamma is disproportionately arguing to are residing alone" –, it asserted, syndication to \setblue{sporting pipelines} would \setblue{betraying} a \setblue{husband's betrothal} in the \setblue{habitation}. \underline{In December, the UK’s \setblue{assemblage} heard} \underline{a \setblue{requisitions} that a \setblue{subscriber} to BSkyB was ‘\setorange{often damaging}’, along with liquor, tobacco and } \underline{gambling.} Conservative MP Alec Shelbrooke was proposing the \setblue{repaying} of benefits and tax credits on a "Welfare Cash Card", in the styling of the Supplemental Nutrition Assistance Program, that could be used to buy only "essentials". \\
\textbf{Model prediction} & \noindent{\setred{damaging}} \\
\hdashline
\textbf{T3 context} & The Daily Mail newspaper reported in 2012 that the UK government's benefits agency was checking claimants' "Sky TV bills to establish if a woman in receipt of benefits as a single mother is wrongly claiming to be living alone" – as, it claimed, subscription to sports channels would betray a man's presence in the household. \underline{In December, the UK’s parliament heard a claim that a subscription to} \underline{BSkyB was ‘\setorange{often damaging}’, along with alcohol, tobacco and gambling.} Conservative MP Alec Shelbrooke was proposing the payments of benefits and tax credits on a "Welfare Cash Card", in the style of the Supplemental Nutrition Assistance Program, that could be used to buy only "essentials". \setgreen{The world it contained to the available than [unk] available sometimes damaged.}\\
\textbf{Model prediction} & often damaging \\
\hdashline
\textbf{TASA context} & The Daily Mail newspaper reported in 2012 that the UK government's profits agency was checking claimants' "Sky tv set throwaways to establish if a woman in receipt of profits as a single mother is wrongly claiming to be living alone" – as, it claimed, subscription to gambols epithelial ducts would betray a man's presence in the household. \underline{In December, the UK’s parliament \setblue{noticed} a} \underline{claim that a subscription to BSkyB was ‘\setorange{often damaging}’, along with alcohol, tobacco and gambling.} Conservative MP Alec Shelbrooke was popping the questioning the requitals of dos goods and tax credits on a "Welfare Cash Card", in the style of the Supplemental Nutrition Assistance Program, that could be used to buy only "essentials". \setgreen{In December, the Bhinmal’s parliament heard a claim that a subscription to BSkyB was ‘meticulously ionateing’, along with alcohol, tobacco and gambling.} \\
\textbf{Model prediction} & \noindent{\setred{meticulously ionateing}} \\
\bottomrule
\end{tabularx}

\begin{tabularx}{\textwidth}{lX}
\toprule
\textbf{Original context} & On May 21, 2013, NFL owners at their spring meetings in Boston voted and awarded the game to Levi's Stadium. \underline{The \$1.2 billion stadium opened in \setorange{2014}}. It is the first Super Bowl held in the San Francisco Bay Area since Super Bowl XIX in 1985, and the first in California since Super Bowl XXXVII took place in San Diego in 2003.  \\
\textbf{Question} & When did Levi's stadium open to the public? \\
\textbf{Answer} & 2014 \\
\hdashline
\textbf{TextFooler context} & On May 21, 2013, NFL owners at their spring meetings in Boston voted and awarded the game to Levi's Stadium. \underline{The \$1.2 \setblue{trillion} stadium opened in \setorange{2014}}. It is the first Super Bowl held in the San Francisco Bay Area since Super Bowl XIX in 1985, and the first in California since Super Bowl XXXVII took place in San Diego in 2003. \\
\textbf{Model prediction} & \noindent{\setred{May 21, 2013}} \\
\hdashline
\textbf{T3 context} & On May 21, 2013, NFL owners at their spring meetings in Boston voted and awarded the game to Levi's Stadium. \underline{The \$1.2 billion stadium opened in \setorange{2014}}. It is the first Super Bowl held in the San Francisco Bay Area since Super Bowl XIX in 1985, and the first in California since Super Bowl XXXVII took place in San Diego in 2003. \setgreen{By by got to to these and 2012.}\\
\textbf{Model prediction} & 2014 \\
\hdashline
\textbf{TASA context} & On May 21, 2013, NFL possessors at their spring runs across in Boston balloted and awarded the game to Levi's Stadium. \underline{The \$1.2 billion stadium opened in \setorange{2014}.} It is the first Super Bowl held in the San Francisco Bay Area since Super Bowl XIX in 1985, and the first in California since Super Bowl XXXVII took place in San Diego in 2003. \setgreen{The \$1.2 billion door opened in 2 June 2013.} \\
\textbf{Model prediction} & \noindent{\setred{May 21, 2013}} \\
\bottomrule
\end{tabularx}
\caption{\label{tab:case2} Adversarial contexts generated by TextFooler, T3, and TASA, compared to the original context on SQuAD 1.1 using BERT as victim model, along with predicted answers by the model. \setorange{Gold answer}, \setblue{perturbed tokens} (i.e. perturbations on answer sentence for TASA),  \setgreen{added distracting sentences} (i.e. DAS for TASA), and \setred{wrong answers} are in different colors. \underline{Underlined sentences} indicate the answer sentences. }
\end{table*}

\end{document}